\documentclass[letterpaper,twocolumn,fleqn]{article} 

\usepackage[pass,letterpaper]{geometry} % Afin d'avoir A US-Letter-sized PDF (demandé par EI).

\usepackage{ist}
\usepackage{amsmath, amssymb, latexsym}
\usepackage{tikz}
\usetikzlibrary{decorations.pathreplacing}
\usetikzlibrary{fadings}
\usepackage{multicol}
\usepackage{url}
\title{\vspace{-2cm}
{\footnotesize \textnormal{\textit{Color Imaging XXIV: Displaying, Processing, Hardcopy, and Appli., IS\&T Int. Symp. on Electronic Imaging, SF, California, USA, 13 - 19 Jan. 2019. \vspace{+1.3cm}\\}}}
A CNN adapted to time series for the classification of Supernovae}

\author{Anthony BRUNEL${}^{1}$, Johanna PASQUET ${}^{2}$, J\'er\^ome PASQUET ${}^{3}$, Nancy RODRIGUEZ ${}^{1}$, \\
Fr\'ed\'eric COMBY ${}^{1}$, Dominique FOUCHEZ ${}^{2}$, Marc CHAUMONT ${}^{1,4}$\\ \\
\small
	$^{1}$ LIRMM, Universit\'e Montpellier, CNRS, France\\
	$^{2}$ CPPM, Aix Marseille Universit\'e, CNRS/IN2P3, France \\
	$^{3}$ IRSTEA, Universit\'e Montpellier, AMIS, Universit\'e Paul Val\'ery, France \\
	$^{4}$ Universit\'e N\^imes, France\\
\small\texttt{\small\{anthony.brunel, nancy.rodriguez, frederic.comby, marc.chaumont\}@lirmm.fr}\\	
\small\texttt{\small\{pasquet, fouchez\}@cppm.in2p3.fr}\\
\small\texttt{\small\{jerome.pasquet\}@univ-montp3.fr}	
}
\date{} % date has an empty field.

\begin{document} 

\maketitle 

\thispagestyle{empty} % prevents the first page to be numbered

%%%%%%%%%%%%%%%%%%%%%%%%%%%%%%%%%%
% Abstract
%%%%%%%%%%%%%%%%%%%%%%%%%%%%%%%%%%

\begin{abstract}
Cosmologists are facing the problem of the analysis of a huge quantity of data when observing the sky. The methods used in cosmology are, for the most of them, relying on astrophysical models, and thus, for the classification, they usually use a machine learning approach in two-steps, which consists in, first, extracting features, and second, using a classifier.  In this paper, we are specifically studying the supernovae phenomenon and especially the binary classification ``I.a supernovae versus not-I.a supernovae''.  We present two Convolutional Neural Networks (CNNs) defeating the current state-of-the-art. The first one is adapted to time series and thus to the treatment of supernovae light-curves. The second one is based on a Siamese CNN and is suited to the nature of data, i.e. their sparsity and their weak quantity (small learning database).
\end{abstract}

%%%%%%%%%%%%%%%%%%%%%%%%%%%%%%%%%%%%
% Overall Document Guidelines: Head
%%%%%%%%%%%%%%%%%%%%%%%%%%%%%%%%%%%%
%%%%%%%%%%%%%%%%%%%%%%%%%%%%%%%%%%%%%%%%%%%%%%%%%%%%%%%%%%%%%%%%%%%%%%%%%%%%%%%%
\section{INTRODUCTION}
Cosmologists are facing the problem of the analysis of a huge quantity of data when observing the sky. As an example, in a very close future (2022), the Large Synoptic Survey Telescope (LSST)\footnote{\url{https://www.lsst.org/}} will produce terabytes of images of the sky per day.  Cosmologists thus need automatic analysis algorithms to alert them when a cosmological phenomenon occurs, such as the explosion of a star (supernova). In order to exploit this huge amount of data, cosmologists and computer-science scientists work in collaboration on classification algorithms, and on benchmarking, for example through international competitions such as the Astronomical Classification Challenge form Kaggle\footnote{\url{https://www.kaggle. com/c/PLAsTiCC-2018}}.
	
In this paper, we are specifically studying the supernovae phenomenon and especially the binary classification "I.a supernovae versus not-I.a supernovae". The type I.a supernovae produce an extremely bright explosion which can be seen at a very far distance. The uniform intrinsic brightness of this kind of star allows to calculate distances and to understand the universe and the dark energy better. Roughly $10^3$ supernovae have been discovered in the history of astronomy. According to \cite{b_lsst1}, the LSST will allow discovering over ten million supernovae during its ten years survey. We thus have to anticipate and find methods that will ease both the processing and analysis of astronomical data.
    
From a practical point of view, when a zone of the sky is observed by a telescope during the night, an image is produced. The light flux from a region of interest in the image - where a phenomenon is occurring - is computed. Cosmologists repeat this calculation every night. Then they can build a time-series giving the light flux in function of the date. In practice, they are making time-series for specific bands: green, red, near-infrared and infrared. 

\begin{figure}[h]
        \centering
        \includegraphics[width=0.5\textwidth]{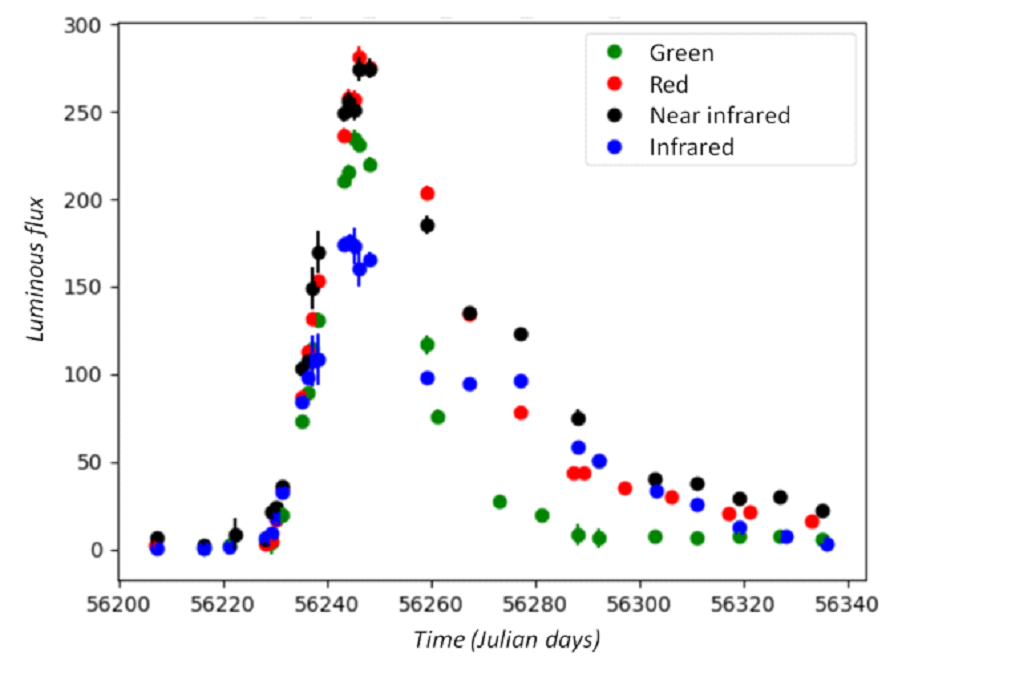}
		\caption{A light-curve example of type I.a supernova.}
        \label{fig:lc}
\end{figure}

In this work, and having in mind what will be produced by the LSST, we use these four times-series as input data. The set of four times-series is named a {\it light-curve}, and its duration is about 100 to 200 days. The Figure \ref{fig:lc} gives an example of a supernova I.a light-curve. light-curves have irregular sampling and are sparse. This sparsity is due to the weather conditions which do not allow a regular observation, and also because the observed zone of the sky is not always the same each night. 

Until the end of 2017, cosmologist methods for classifying a light-curve as a supernova type I.a or not type I.a, are mostly relying on the use of astrophysical models. They usually use a two-steps machine learning approach: feature extraction and classification. Feature extraction is thus a crucial step, but even if features are chosen by experts, they can be incomplete or sub-optimal for the classification step. In this paper, we present two Convolutional Neural Networks (CNNs) to replace the usual two-steps machine learning approaches. The first CNN is adapted to time series and thus to the treatment of supernovae light-curves. The second one is based on a Siamese CNN \cite{b_siamese} and is adapted to the nature of data, i.e. their sparsity and their weak quantity (small learning database). We benchmarked our CNNs to two well established state-of-the-art methods, the first one by  Lochner {\it et al.} \cite{b_lochner}, and the second one using {\it FATS features} of Nun {\it et al.} \cite{b_fats}. Additionally, we also evaluated our best CNN to a recurrent neural network (RNN) proposed by Charnock and Moss \cite{b_rnn1}, which is also adapted to time series. Confronted by the strong sparsity of the data, our CNN performs better. Note that we took care to compare the algorithms in fair conditions, at low regime (small learning database), and also with hard condition close to real conditions, by having more than 70\% of unavailable data in each time series. 

This paper is organized as follows. In Section \ref{part2} we present the two state-of-the-art methods of supernovae classification. These machine learning algorithms use different feature descriptors; one relies on supernovae models and the other one on times series. We also present a deep learning method using a recurrent neural network \cite{b_rnn2}. In Section \ref{part3} our CNN and Siamese CNN architecture are presented. Then, in Section \ref{part4} we detail the databases, the experiments setup, the parameters settings, and analyze results. In Section \ref{part5} we conclude and give possible extensions of this work.

\section{Existing methods}\label{part2}
In this section, we present a small survey of recent machine learning-based methods for supernovae classification. The first method, proposed by Lochner {\it et al.}~\cite{b_lochner}, uses {\it SALT2} features \cite{b2}, and was considered as the state-of-the-art until the end of 2017. The second method relies on the use of a library named ``Feature Analysis for Time Series'' (FATS) whose features are dedicated to light-curves analysis \cite{b_fats}. Finally, we also present the work of Charnock and Moss \cite{b_rnn1} which uses a recurrent neural network.

\subsection{Boosted decision trees using SALT2 features}\label{BDTs}

In \cite{b_lochner}, the authors compare the results of different classifiers and features. They found that the  boosted decision trees (BDT) with features of ``Spectral Adaptive Light-curve Template features 2'' (SALT2) model provided the best performance. SALT2 is the most commonly used model for type I.a supernovae. It relies on the following equation for computation of the light-curve's flux in a given band:
        	\begin{equation} 
            \begin{split}
				F(t,\lambda) = x_{0}&\times[M_{0}(t,\lambda) + x_{1}M_{1}(t,\lambda) + ... + M_{k}(t,\lambda)]\\
				& \times\exp[c \times CL(\lambda)],			\label{SALT2eq}    
            \end{split}
           	\end{equation}
where $t$ is the time since the maximum luminosity in B-band  (the phase), $\lambda$ is the wavelength, $M_{0}(t,\lambda)$ is the average spectral sequence, $M_{k}(t,\lambda)$ for $k>0$ are higher order components that describe the variability of supernovae I.a. $CL(\lambda)$ represents the average color correction law. For each supernova, the redshift (variation of the wavelength due to the expansion of the universe) $z$ is also used. The machine learning algorithm uses five parameters to describe the supernovae: $z$, $t_{0}$ (the time of peak brightness in the B-band), the normalization term $x_{0}$, $x_{1}$ that describes the shape of the light-curve, and $c$ the color at the maximum luminosity in the B-band. Then, BDT are used to classify the supernovae. 
BDT are machine learning classifiers using multiple decision trees to construct a model. It associates input features to output classes. Multiple decision trees are built on slightly different subsets of data, and the resulting classifications are averaged to provide robustness. Instead of using bagging like in random forest (RF) \cite{b_RF}, which selects subsets of data with random replacement, a BDT use boosting such that for each iteration the same data-set is used but with an increase of the weights of incorrectly classified examples, which allows subsequent classifiers to focus on difficult cases. Even if Lochner et al. \cite{b_lochner} found that RF and BDT gave almost the same classification results, BDT were usually faster than RF. Moreover, BDT are considered as robust classifiers because of the averaging process \cite{b_Ensemble}. %However, they are sometimes computationally expansive.%retiré suite à la remarque de Johana 

\subsection{Boosted decision trees using FATS library}\label{fats}

In Lochner {\it et al.} \cite{b_lochner}, in addition to SALT2, several feature extraction methods are compared. Among them, there is a method that uses no prior knowledge on supernovae light-curves models, which makes it a generic approach. Nevertheless, similarly to what was observed in \cite{b_lochner}, our experimental results show that this wavelet-based method is less efficient than the approach of Lochner {\it et al.}. We thus choose another generic approach based on the FATS library \footnote{FATS is a python library, and it can be found on GitHub https://github.com/isadoranun/FATS} \cite{b_fats}, which requires no previous supernovae knowledge and allow to extract many more features than SALT2. FATS (Feature Analysis for Time Series) is a python library which includes a set of features dedicated to time series analysis. FATS contains relevant information for the classification of astrophysical objects, such as the color which is the difference of the flux between two separate bands, the skewness, the mean, ... We then use BDT (presented in \ref{BDTs}) to perform the classification.

\subsection{Recurrent neural network for supernovae classification}\label{ssec::rnn}
Recurrent neural networks (RNN) are a class of deep learning method which exhibit a temporal dynamic behavior for a time sequence \cite{b_rnn2}. RNN are used for tasks such as speech recognition \cite{b_rnn3} and language translation \cite{b_rnn4}. %They have also been used for supernovae light-curves classification in the work of Charnock and Moss \cite{b_rnn1}. Two kinds of RNNs exist: unidirectional and bidirectional RNNs. %Unidirectional mean that the information flow goes in one direction; bidirectional means that the  information is able to propagate forwards and backwards. 
In \cite{b_rnn1}, the authors present a supernovae type I.a versus not-I.a classification algorithm using RNN with Long Short-Term Memory (LSTM) cells \cite{b_lstm}. They used LSTM cells because more traditional RNN are unable to manage long-term information. By using the gating process, the LSTM architecture overcomes the vanishing gradient problem, which means that it is more able to learn on long time series while taking better into account the past values. Each LSTM cell is composed of three gates. The first one represents the forget gate that allows the network to remove the information transmitted by the previous cell. The second corresponds to the input gate which processes the input information at a given time. The last gate merges the information from the input gate and the output gate to feed the next cell of the LSTM network with a piece of new information.
In \cite{b_rnn1}, the authors tested multiple RNN LSTM architectures with different numbers of LSTM cells per hidden layer. The RNN architecture that gave the best results is unidirectional with two hidden layer and sixteen LSTM cells per layer. %Their RNN take as input vector to each sequential step: the time (in days since the first observation), the flux in each of the  4  bands,  flux  errors  in  each  of  the  4  band and some other astronomical parameter like RA and Dec, dust extinction; and host photo-z (redshift) if relevant. Also they do data augmentation for the missing value for each observation.

\section{NETWORK ARCHITECTURE}

In the section, we present the most important characteristics of our convolutional neural network (CNN) and our Siamese network.\\

\subsection{Convolutional neural network} \label{part3}
In this section, we present our convolutional neural network (CNN). First, we explain its architecture, and then we detail some of the most important elements.\\

\subsubsection{Network architecture}

\begin{figure*}[t!]
        \centering
        \includegraphics[width=1.0\textwidth]{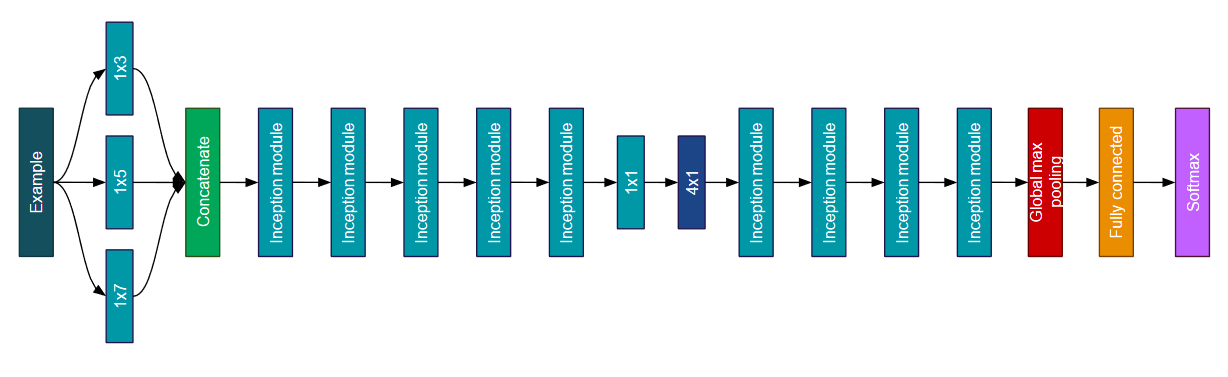}
 		\caption{Our convolutional neural network. There is three parallel temporal convolution followed by a depth concatenation. Then, there are five inception modules, whose two of them have strides of two for the width reduction. Next, there is the color convolution layer with a $1\times1$ filter, followed by a $4\times1$ filter. After, there are four inception modules, whose two of them have strides of two. Finally, there is the global max pooling, a 1024 fully connected layer, and the softmax layer which give the classification result.}
        \label{fig:net}
\end{figure*}

Our CNN is shown in Figure \ref{fig:net}. The source code can be downloaded at \url{https://github.com/Anzzy30/SupernovaeClassification}. Our CNN takes as input a light-curve which is represented by a matrix of dimension $4\times T$, where 4 stands for the height and T for the width. There are 4 rows (one for the green, one the red, one the near-infrared, and one for the infrared bands), i.e. 4 time-series, and the duration T of the light-curve is variable. The CNN contains eleven layers, and most of those layers are "inception" modules (described below), where the filters are modified to be 1D temporal filters (performed along the rows of the input). Poolings are also modified to be 1D. A time series is indeed usually represented as a 1D array, and this ensures the extraction of temporal information independently for each time series. 1D convolutions capture the input signals evolution over the time \cite{b_tconvolution}. Layers 2, 5, 9 and 10 are inception modules with 1D convolutions and a stride of two pixels. This enables to divide the input width by two after each inception block.

We also introduce a depth-wise convolution (a 1D convolution in the depth axis, i.e. a convolution in the color axis) in the 7th layer to take into account the correlation between the bands. This color-convolution is done without padding, over the four columns in order to merge the time series and combine the information provided by each time-series. This fusion process is important for a more efficient separability between the two classes supernovae I.a and not-I.a \cite{b_color}. We indeed observed an increase in the accuracy of the classification by using this color-convolution. 

Each convolution operation is followed by a ReLU (Rectified Linear Units) activation function. At the end of last convolution layer, we perform a global max pooling as described in subsection \ref{gmp}.
This "global max pooling" allows treating light-curves of any duration. Then, a fully connected network with 1024 neurons is used for the classification part. The network's output is finally provided by a softmax prediction between the two classes I.a supernovae and not-I.a supernovae.\\

\subsubsection{Inception module}\label{module} 
\begin{figure}[h]
        \centering
        \includegraphics[width=0.5\textwidth]{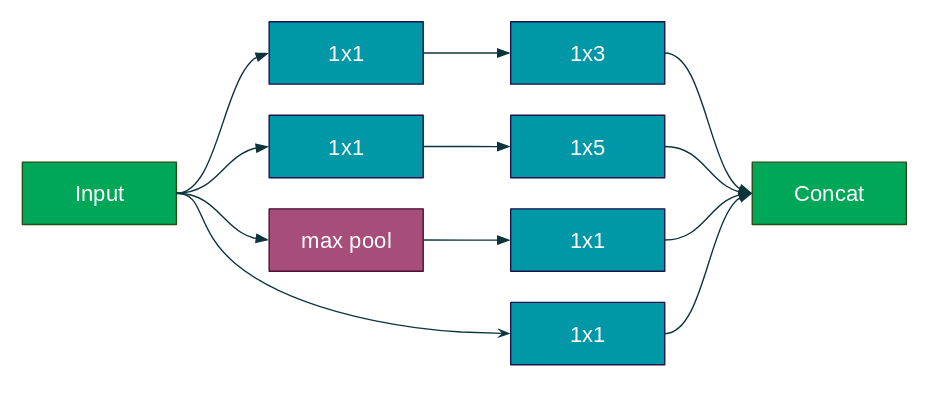}
		\caption{Our 1D inception module.}
        \label{fig:module}
    \end{figure}

Our network is mainly built by a succession of inception modules \cite{b_googlenet}. These modules have been adapted to work with time series, i.e. 1D convolution. It is built with $1 \times 1$, $1 \times 3$ and $1 \times 5$ convolution layers and max pooling. It allows extracting information at multiple resolutions. $1\times3$ and $1\times5$ convolutions are preceded by $1\times1$ convolution to reduce the dimensionality and add non-linearity. In our network, each convolution layer is followed by a ReLU activation function \cite{b_relu}. The Figure \ref{fig:module} describes our inception module.
	\\
    
\subsubsection{Global max pooling}\label{gmp} 

\begin{figure}[h]
        \centering
        \includegraphics[width=0.5\textwidth]{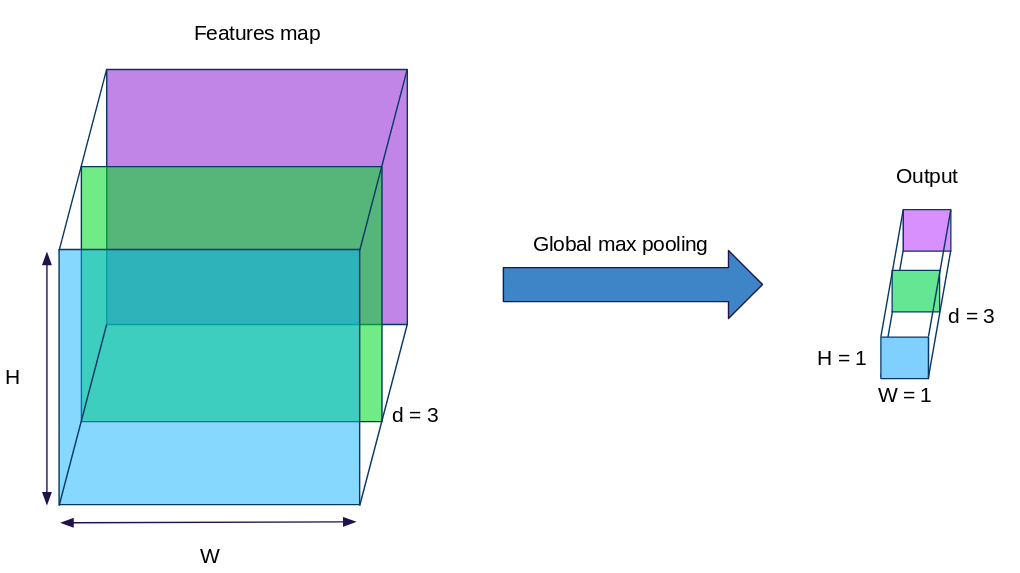}
		\caption{Scheme of the global max pooling mechanism.
        Global max pooling perform an operation that take the maximum per features map and produce an 1D vector with known size (number of features map)}
        \label{fig:gmp}
    \end{figure}

The time series may have a variable duration, i.e. a variable number of columns for the input matrices. A classical CNN is generally built with a fully connected part, this means with a fixed number of neurons, which impose a fixed dimension for the inputs. Thus, for a classical CNN, a variable length of time series generate feature maps of variable lengths, which cannot be processed by the fully connected layer. To overcome this limitation, we incorporate a global max pooling layer (described in Figure \ref{fig:gmp}) after the last convolution and just before the fully connected layer. This force the features to have a fixed dimension before entering the fully connected layer. Our network can thus classify light-curves of any duration. Another interesting effect of this measure is that during the learning phase, we can now sometimes crop the light-curves from 40\% to 80\% of their original duration. This allows a data-augmentation, reduces the over-fitting, and increases the generalization abilities.

\subsection{Siamese network}\label{ssec:siamese}

The second method proposed in this paper is a Siamese network. This type of neural network was introduced in the 1990s by Bromley et al. \cite{b_siamese} and contains at least two sub-networks with shared weights. Each sub-network produces an n-dimensional feature vector. Afterwards, we can compute different metrics like L2-norm between vectors. The learning process for this method is to bring closer, in the features space, elements that have the same label and drive away elements with a different label.

With the Siamese network we search for a solution to the sparsity problem which is extremely present in the data (See Fig. \ref{fig:lc}). We propose a loss function with the triplet loss presented in \cite{b_triplet}, and we also propose an adaptation of the triplet loss. First, each triplet is chosen online (online triplet mining) which means that the useful triplets are computed on the fly. Online triplet mining was introduced in Facenet\footnote{http://bamos.github.io/2016/01/19/openface-0.2.0/}. Online triplet mining is more efficient than offline regarding computation time and performance. 

The loss function is described by Equations \ref{eq1} and \ref{eq2}. First, we compute the classic triplet loss that allows the network to bring closer elements with the same label. The triplet loss is defined by the Equation \ref{eq1},
\begin{equation} 
				L_{triplet}(a,p,n) = \max(d(a,p) - d(a,n) + margin, 0),\label{eq1}    
\end{equation}
with $d$ the L2 norm function which takes two vectors as input, $a\in\mathbb{R}^{m}$ a feature vector of dimension $m$, named the anchor, $p\in\mathbb{R}^{m}$ a feature vector named the positive example (it has the same label than the anchor), $n\in\mathbb{R}^{m}$ a feature vector associated to the negative example (it has a different label as that of the anchor), and the $margin\in\mathbb{R}$. This triplet loss, when minimized, has the effect of pushing the negative examples at a distance of the anchor greater than the margin plus the distance of the anchor to the positive.
            
During the optimization process, the triplet examples whose loss is greater than zero are the only triplets useful for the minimization. The minimization is thus done on the arithmetic mean of the useful triplets; see Equation \ref{eq2}:
\begin{equation} 
				L_{final\_triplet} = \frac{1}{N-1}\sum_{j = 0}^{N-1}L_{triplet_{j}}, \label{eq2}
\end{equation}
with $N$ the number of useful triplet $L_{triplet_{j}}$.
            
In order to better take into account the sparsity of light-curves, in addition to the triplet loss function, we propose an adaptation of the triplet loss described by Equation \ref{eq3}. The goal of this additional loss function is to amalgamate ($=$ force an identical feature representation) the anchor features vector with the feature vector obtained by a sub-sampling of the anchor light-curve, while keeping a minimum distance with negative examples.
\begin{equation} 
				L_{triplet^\prime}(a,a^\prime,n) = d(a,a^\prime) + \max(0, margin^\prime - d(a,n)), \label{eq3}
\end{equation} 
where $a$ $\in\mathbb{R}^{m}$ is a feature vector associated to the anchor example, $a^\prime$ $\in\mathbb{R}^{m}$ is a feature vector of the sub-sampled anchor light-curve, $n$ $\in\mathbb{R}^{m}$ is a feature vector associated with the negative example (which is an example different from the anchor), and the $margin'\in\mathbb{R}$.

Once again, during the optimization process, the triplet examples whose loss is greater than zero are the only triplets useful for the minimization. The minimization is thus done on the arithmetic mean of the useful triplets; see Equation \ref{eq4}:
\begin{equation} 
				L_{final\_triplet^\prime} = \frac{1}{N-1}\sum_{j = 0}^{N-1}L_{triplet^\prime}. \label{eq4}
\end{equation} 
            
\begin{figure}[h]
        \centering
        \includegraphics[width=0.5\textwidth]{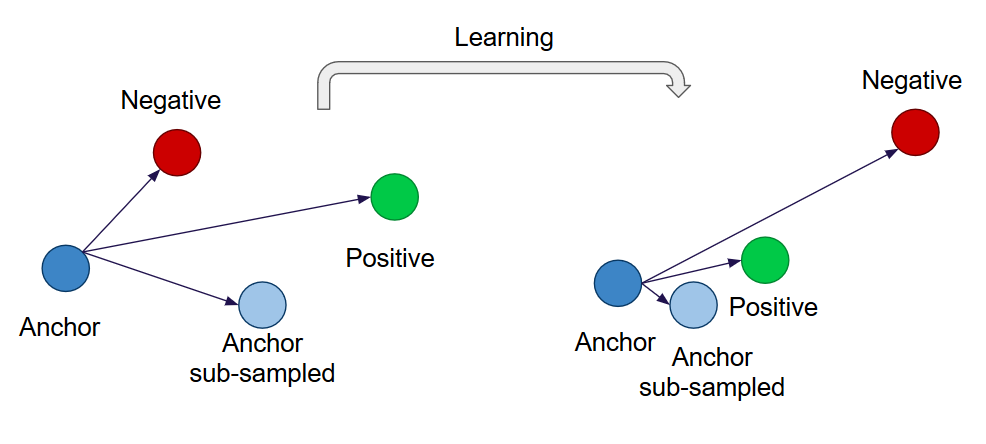}
		\caption{Scheme of the learning process of our adapted triplet loss function; See the Equation \ref{eq5}. It aims (i) to amalgamate the Anchor feature vector with the Anchor feature vector issued from the sub-sampling of the same light-curve, (ii) to bring closer, without amalgamation, the Anchor vector and the Positive vector, and (iii) to keep distant the Anchor vector and the Negative vector. For example: {\it Anchor} is a supernova I.a, {\it Anchor sub-sampled} is the version of the Anchor resultant of a sub-sampling of same the light-curve, {\it Positive} is another supernova I.a, and {\it Negative} is a supernova not-I.a.}
        \label{fig:tripletlearning}
\end{figure}

The final loss (Equation \ref{eq5}) is the sum of the two previous loss functions described by Equations \ref{eq2} and \ref{eq4}. Figure \ref{fig:tripletlearning} illustrates the learning process of the Siamese network.
\begin{equation} 
				L_{final} = L_{final\_triplet} + L_{final\_triplet^\prime}.
              \label{eq5}
\end{equation} 
       
To compute the features vectors of each example, we use two networks with shared weights and the same architecture as the convolutional neural network defined in section \ref{part3} but without the fully connected layer. %Thus, one network computes the feature vectors of the batch example and the other computes the feature vectors of the sub-sampled example. MC: Le 16/11/2018 COmprend pas cette phrase
Once the Siamese network converges, we train a fully connected layer of 1024 neurons followed by a softmax layer. The input is the feature vector (output of the Siamese network), and the outputs are the two classes to predict (type I.a or not-I.a).

\section{EXPERIMENTS}\label{part4}

In this section, we present the database used for the experiments, the parameter settings, the setup for the experiments, and finally, our results.

\subsection{Base}\label{BASE}

\subsubsection{First base}  We simulate with the software SNANA \cite{b_snana} light-curves of different types of supernovae.  SNANA is an analysis package for supernovae light-curves that contains a simulation, a light-curve fitter and a cosmology fitter. It takes into account actual survey conditions and so generates realistic light-curves by using the measured observing conditions at each survey epoch and sky location. Each light-curve is composed of 4 time-series. Each time series contains the light flux measure in specific bands taken each day during a certain period of time (around 100 days). The bands are obtained using color band-pass whose filters are green, red, near-infrared, and infrared. In the first experiment, we used 5000 light-curves with 2500 supernovae of type I.a and 2500 of type not-I.a. For the deep learning input, each data is day-sampled and stored in a matrix. Each cell has a specific value of flux. If a value is missing, then we fill it up by zero values. As light-curves of supernovae are very sparse (due to missing values), the matrix contains more than 70\% of zero values. We will note this database B1.\\

\subsubsection{Second base} The second database is available on GitHub\footnote{https://github.com/adammoss/supernovae} and contains 21319 light-curves with 5088 I.a supernovae and 16231 not-I.a supernovae. This database is from the Supernova Photometric Classification Challenge \cite{Kessler_SPCC_2010}. We used this database to compare our CNN (our first architecture; See Section \ref{part3}) to the results obtained by the RNN LSTM of \cite{b_rnn1} (See Section \ref{ssec::rnn}). We will note this database B2.\\

\subsubsection{Data augmentation} For the convolutional neural networks, we used artificial data augmentation. This method allows the networks to get a better representative set for the learning process. Each light-curve gets a chance to be altered during each epoch. The alteration will crop the light-curve and takes a random fraction between 0.4 and 0.8 of the light-curves. Then the network is fed with this new representation. This strategy reduces over-fitting and slightly increases the classification performances.

\subsection{Parameter settings}
For the machine learning methods, we used the algorithm available on Lochner's GitHub\footnote{https://github.com/MichelleLochner/supernova-machine} with boosted decision trees, as presented in \cite{b_lochner}.

We set the number of iteration at 4500 for the convolutional neural network and 9000 for the Siamese network. We set the dropout to $0.4$ for the fully connected layer to reduce over-fitting. The learning rate for the 2 deep learning methods varies between $1\times 10^{-2}$ and $5\times 10^{-4}$
%$\num{1e-2}$ and $\num{5e-4} $ with exponential decay. For the two methods we used Adam
with exponential decay. For the two methods, we used Adam optimizer \cite{b_adam} on a cross entropy loss function. Network's weights are initialized with Xavier uniform initializer algorithm \cite{b_xavier} and the batch size is set to 128. 

\subsection{Experiments setup}

We used tensorflow and python to develop our deep learning methods. The training phase is performed with an NVIDIA GTX 1080.

To compare the two methods relying on a two-steps machine learning approach with our convolutional neural network and our Siamese network, we used the Base B1 with k-folds cross-validation with k=4 and 5000 light-curves. The database is thus randomly partitioned into 4 equal sized sub-samples. A single sub-sample is kept for testing the model, and the remaining 3 sub-samples are used as training data. We repeat the process 4 times, with each of the k sub-samples used exactly once as the testing data. It means that each model is trained on only 3750 light-curves and tested on the 1250 other.

To confront our convolutional neural network against the recurrent neural network \cite{b_rnn1}, we followed the same strategy as presented in \cite{b_rnn1} with the database B2. We thus trained our CNN model (our first architecture; See Section \ref{part3}) on a base composed of 5330 supernovae light-curves took randomly from the database B2. The remaining elements in B2 (15989) are used to evaluate the network. We repeat this process five times to compute the arithmetic mean and the standard deviation of the results. Note that the number of training data is chosen to be close to the training set size used with the database B1.

\subsection{Comparison metrics}

We use two metrics to compare the different methods. The first is the accuracy defined as the ratio of the number of correct predictions over the total number of predictions. The second one is the Area Under the ROC Curve (AUC). The ROC curve represents the True Positives Rate (TPR) versus False Positives Rate (FPR) when the probability threshold is moved from 0 to 1. TPR and FPR are given by equations \ref{TPR} and \ref{FPR}.
\begin{multicols}{2}
  \begin{equation}\label{TPR}
    TPR =  \frac{TP}{TP + FN} 
  \end{equation}\break
  \begin{equation}\label{FPR}
    FPR = \frac{FP}{FP + TN}, 
  \end{equation}
\end{multicols}
where TP are the true positives, FP the false positives, TN the true negatives and FN the false negatives. %For multiple class problem (two in our case), AUC is calculated by averaging AUC for each class.
\subsection{Results and discussion}
    
\subsubsection{Database B1: Comparisons of our CNNs and our Siamese network, to SALT2, and FATS}   
\begin{figure}[h]
    \centering
    \includegraphics[width=0.5\textwidth]{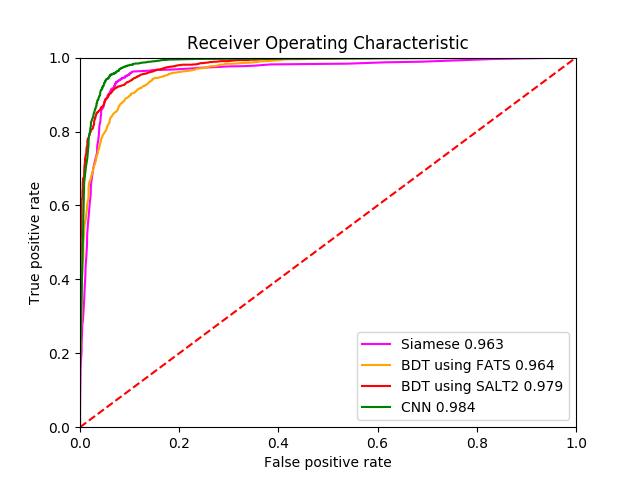}
	\caption{ROC curves of the four methods: Siamese, BDT using FATS, BDT using SALT2, and our CNN.}
    \label{fig:roc}
\end{figure}

In our first experiment with the database B1 we confront our CNN and our Siamese network to BDT + SALT2 \cite{b_lochner}, and BDT + FATS \cite{b_fats}. The Figure \ref{fig:roc} shows the obtained ROC curves. The convolutional neural network gets the best performances with an AUC equal to $98.4\%$ followed by the BDT with SALT2 features which obtains $97.9\%$. The Siamese network and the BDT using FATS obtain respectively an AUC equal to $96.3\% $ and $96.4\%$.

Another interesting metric is the accuracy. The CNN got the best accuracy of all 4 methods with $94.6\%$ (See Table \ref{sec:table}). Siamese network comes with $93.0\%$ that outperforms BDT using SALT2 which obtains $92.3\%$. BDT using FATS get the lowest accuracy with $90.1\%$.

\begin{table}[h]
\centering
\begin{tabular}{l*{6}{c}r}
Model & Training set & AUC & Accuracy \\
\hline 
CNN & 3750 & $\textbf{0.984}$ & $\textbf{94.6}$\\
SALT2 & 3750 & 0.979 & 92.3\\
FATS & 3750 &0.964 & 90.1\\
Siamese & 3750 & 0.963 & 93.0\\
\end{tabular}
\caption{Table1. AUC and accuracy for the four methods (Siamese, BDT using FATS, BDT using SALT2, and our CNN) on B1 database.}
\label{sec:table}
\end{table}

The state-of-the-art in two-steps machine learning methods used for supernovae classification \cite{b_lochner}, \cite{b_fats},  perform well with a small learning set (i.e. less than 1 000 learning example), but with the above results, we can clearly say that CNNs are a promising solution for the binary classification of super-novae type I.a versus not-I.a. Indeed, our CNN performs better than two-steps machine learning methods which are based on hand-crafted features, and this with a small number of examples (3 750). A set of 3 750 examples is indeed considered as a very small database for a learning of a binary classification by deep learning methods. 

The Siamese network gets good results but is harder to train because of the shared weights between its two networks. Even if its results are not as good as our CNN, we think the results can be improved with a better loss function. Adding some data augmentation, as we do with our CNN, may also improve the performances.

Additionally, we observed that with the increase in the size of the learning set (i.e. with more time series) the performances of our CNN are increasing. Another advantage of our method is that it learns the filters value directly to extract the more relevant features. There is thus no need of prior knowledge as in SALT2 \cite{b_lochner} or FATS \cite{b_fats}.

When working with a more complex architecture, we have shown that in a more realistic and more complicated situation from a cosmological point of view (extremely small database, a mismatch between learning set and test set), that Deep Learning approaches surpass all the previous known approaches \cite{Pasquet_Pelican_2019}.\\

\subsubsection{Database B2: Comparisons of our CNN with an RNN LSTM}   
In the second experiment with the database B2, we confront our CNN against the recurrent neural network proposed in \cite{b_rnn1}. The input vector given to the RNN contains the day, the light flux in each band, and the redshift extra-information. The RNN thus uses a little bit more information compared to our CNN. In this experiment, 5330 supernovae light-curves are used for the learning, whereas in the first experiment 3750 light-curves were used. We choose 5330 light-curves (a fraction of 25\% of the database) because it is close to the size used in our first experiment. In this experiment, the proportion of I.a and not-I.a is around 1/4, 3/4 whereas in the first experiment with B1 it was 1/2, 1/2. For our CNN the data-augmentation is done by cropping, and it artificially increases the learning data-set by 2 or 3. For the RNN, there is a random interpolation which artificially increases the learning data-set by 5. The data-augmentation for this experiment improves the results for both of the approaches of roughly 1\% (with a smaller increase for our CNN). The comparison should then be considered more or less fair, with nevertheless a small disadvantage for our CNN.

\begin{table}[h]
\centering
\begin{tabular}{l*{6}{c}r}
Model & Training set & AUC & Accuracy \\
\hline 
CNN & 5330 & $\textbf{0.983} \pm \textbf{0.0004}$ & $\textbf{94.1} \pm \textbf{0.09}$   \\
RNN & 5330 &$0.975 \pm 0.003$& $92.9 \pm 0.6$   \\
\end{tabular}
\caption{Table 2. AUC and accuracy for our CNN and RNN LSTM on B2 database.}\label{rnnresult}
\end{table}

The results (see Table \ref{rnnresult}) show that our CNN is superior to the RNN LSTM of \cite{b_rnn1} with 98.3\% of AUC against 97.5\%, and 94.1\% of accuracy against 92.9\%. Also, note that the standard deviation of the five runs is smaller for our CNNs. In \cite{b_rnn1} multiple experiments with a different fraction of the database are presented. Taking more data, for example, a fraction of 50\% of the database, provides better results for both our CNN and the RNN, and the gap between our CNN and the RNN remains the same. 

The recurrent neural networks are a great technique to treat time series as it allows to extract dynamic temporal behavior for a time sequence. However, in this work, we showed that our CNN is better. This is maybe due to the trend of over-fitting that is more present in the RNN. The various information of different nature, given to the RNN (day, light flux in each band, redshift), can also make information treatment harder. 

Note that when taking less data, for example, a fraction of 5\% ($=$ 1065 light-curves) of the database, SALT2 method  \cite{b_lochner} gets the best results. When the size of the learning set is too small (less than 1000 light-curves), deep learning approaches are suffering from an insufficient number of examples. Measures can nevertheless be taken such as data-augmentation by noise addition, use of transfer learning or curriculum learning, use of cosmological parameters such as the red-shift, use of ensembles, use of multiple classes, etc. (see for example our paper under revision \cite{Pasquet_Pelican_2019}).

\section{CONCLUSION}\label{part5}
The field of Cosmology is facing great challenges in order to be able to analyze a huge amount of data. One of those challenges is to be able to automatically detect I.a supernovae from not-I.a supernovae. In this paper, we propose a new CNN, adapted to time series (light-curve), that can defeat the state-of-the-art. We also compared our CNN to a Siamese approach and an RNN LSTM. Our CNNs gives the better results, nevertheless, the main conclusion is that all those deep learning approaches are very promising. 

We observed strong efficiency improvement of our CNN when the learning database increases in size. We are also aware that using more clever data-enrichment could boost the efficiency of the different deep learning approaches, and especially our CNN, without substantial additional processing. 

We believe that this publication will reinforce the use of deep learning in the cosmology and astronomy fields. We put all the necessary files for the deployment of our CNN on an open-access GitHub (\url{https://github.com/Anzzy30/SupernovaeClassification}), thus giving strong visibility of our work to the cosmology community.

%%%%%%%%%%%%%%%%%%%%%%%%%%%%%%%%%%
% Bibliography
%%%%%%%%%%%%%%%%%%%%%%%%%%%%%%%%%%

\small
\bibliographystyle{unsrt}
\bibliography{biblio}

%\section{Acknowledgments} 
%add the acknowledgement section here


\vfill\pagebreak
\begin{biography}
%Please submit a brief biographical sketch of no more than 75 words. 
%Include relevant professional and educational information as shown 
%in the example below.
%
%Jane Doe received her BS in physics from the University of Nevada (1977) 
%and her PhD in applied physics from Columbia University (1983). Since 
%then she has worked in the Research and Technology Division at Xerox 
%in Webster, NY. Her work has focused on the development of toner adhesion 
%and transport issues. She is on the Board of  IS\&T and a member of APS 
%and SPIE.

Anthony BRUNEL received his Master's degree in Computer Science in 2018 from the University of Montpellier, France. He is currently working toward the PhD degree in the LIRMM laboratory of Montpellier. His research interests are localization for vision purpose.

Johanna PASQUET received her PhD in 2016 from the University of Montpellier, France on the search of the baryonic dark matter in our Galaxy. She is currently working toward a postdoc in the CPPM laboratory of Marseille on a deep learning approach to observational cosmology with Supernovae. Her research interests are classification with Deep Learning, photometric redshifts and supernovae cosmology.

J\'er\^ome PASQUET is Assistant Professor in computer science at the Universit\'e Paul Val\'ery Montpellier. He conducted a PhD thesis under the supervision of Marc Chaumont and defended it in November 2016 at LIRMM laboratory of Montpellier. His research interests are deep learning representation applied to the objects detection, steganalysis problems and time series classification. 

Nancy RODRIGUEZ is Assistant Professor in Computer Science at the University of Montpellier, France, since 2006. She holds a bachelor's degree in physics and earned her master's degree in computer engineering and physics from Los Andes University, Colombia. She received his PhD in Computer Science from Paul Sabatier University of Toulouse, France. Her research interests include Virtual and Augmented reality, Visualization and Interaction.

Fr\'ed\'eric COMBY received his M.Sc. degree in automatic and microelectronic systems in 1998, and the PhD degree in automatic and signal processing in 2001 from the University of Montpellier, France. He joined the ICAR Team (image and interaction), in the LIRMM (Laboratory of Informatics, Robotics, and Microelectronics of Montpellier) as Assistant Professor in 2003. His current research topics include image processing, vision and multimedia security.

Dominique FOUCHEZ is Research Director at CPPM (Center of Particle Physics of Marseilles) CNRS, AMU (Aix Marseilles University), France. He received his 
PhD in particle physics in 1993 from Aix Marseilles University. His main research interests are fundamental physics and cosmology. 

Marc CHAUMONT is Associate Professor (HDR Hors-Classe) accredited to supervise research since 2013, at the LIRMM laboratory (Laboratory of Computer Science, Robotics and Microelectronic), University of Montpellier and University of N\^imes, in France. He joined the LIRMM in September 2005. He received his PhD in Computer Sciences at IRISA Rennes in 2003. His main research interests are in steganography, steganalysis, digital image forensic, and objects detection with Deep Learning. 
\end{biography}

\end{document}